\documentclass{article}

\usepackage[preprint]{neurips_2025}

\usepackage{graphicx}
\usepackage{booktabs}
\usepackage{multirow}

\usepackage[utf8]{inputenc} 
\usepackage[T1]{fontenc}    
\usepackage{url}            
\usepackage{amsfonts}       
\usepackage{nicefrac}       
\usepackage{microtype}      
\usepackage{xcolor}         
\usepackage{amsmath}
\usepackage{makecell}
\usepackage{subcaption}
\usepackage{arydshln}
\usepackage{wrapfig}
\usepackage{subfig}
\usepackage[ruled,vlined]{algorithm2e}
\usepackage{bm}
\newcommand{\name}{\textbf{LLaVA-CMoE}}

\title{LLaVA-CMoE: Towards Continual Mixture of Experts for Large Vision-Language Models}

\author{
 \textbf{Hengyuan Zhao\textsuperscript{1}\thanks{Equal Contribution}},
 \textbf{Ziqin Wang\textsuperscript{1}\footnotemark[1]},
 \textbf{Qixin Sun\textsuperscript{1}\footnotemark[1]},
 \textbf{Kaiyou Song},
 \textbf{Yilin Li},\\
 \textbf{Xiaolin Hu\textsuperscript{2}\thanks{Corresponding author}},
 \textbf{Qingpei Guo\footnotemark[2]},
 \textbf{Si Liu\textsuperscript{1}\footnotemark[2]},
\\
 \textsuperscript{1}School of Artificial Intelligence, Beihang University \\
\textsuperscript{2}Tsinghua University \\
}

\begin{document}
\maketitle

\setlength{\intextsep}{4pt plus 1pt minus 2pt}
\setlength{\textfloatsep}{4pt plus 1pt minus 2pt}

\begin{abstract}

Mixture of Experts (MoE) architectures have recently advanced the scalability and adaptability of large language models (LLMs) for continual multimodal learning. However, efficiently extending these models to accommodate sequential tasks remains challenging. As new tasks arrive, naive model expansion leads to rapid parameter growth, while modifying shared routing components often causes catastrophic forgetting, undermining previously learned knowledge.
To address these issues, we propose \name, a continual learning framework for LLMs that requires no replay data of previous tasks and ensures both parameter efficiency and robust knowledge retention. Our approach introduces a Probe-Guided Knowledge Extension mechanism, which uses probe experts to dynamically determine when and where new experts should be added, enabling adaptive and minimal parameter expansion tailored to task complexity. Furthermore, we present a Probabilistic Task Locator that assigns each task a dedicated, lightweight router. To handle the practical issue that task labels are unknown during inference, we leverage a VAE-based reconstruction strategy to identify the most suitable router by matching input distributions, allowing automatic and accurate expert allocation. This design mitigates routing conflicts and catastrophic forgetting, enabling robust continual learning without explicit task labels.
Extensive experiments on the CoIN benchmark, covering eight diverse VQA tasks, demonstrate that LLaVA-CMoE delivers strong continual learning performance with a compact model size, significantly reducing forgetting and parameter overhead compared to prior methods. These results showcase the effectiveness and scalability of our approach for parameter-efficient continual learning in large language models. Our code will be open-sourced soon.
\end{abstract} 
\section{Introduction}
Multimodal large language models (MLLMs)~\cite{caffagni-etal-2024-revolution,10386743,Li2023BLIP2BL,Radford2021LearningTV,liu2023visual} have demonstrated remarkable capabilities in understanding and generation across multiple modalities. 
Following large-scale pre-training, these models typically require further adaptation to perform effectively on specific downstream tasks. 
To enable efficient adaptation without necessitating full model retraining, Parameter-Efficient Fine-Tuning (PEFT) methods~\cite{abou2024parameter,han2024parameterefficientfinetuninglargemodels,hu2022lora,houlsby2019parameter,liu-etal-2022-p,liu2021p,edalati2022krona,zhang2022adaptive,chen2022empowering} are widely adopted. These methods ensure computational efficiency and scalability while achieving competitive task-specific performance.

%
Currently, MLLM training is commonly conceptualized as a multi-task learning (MTL) paradigm where all tasks' data are simultaneously accessible.
\begin{figure}[tp]
    \centering
    \includegraphics[width=1\linewidth]{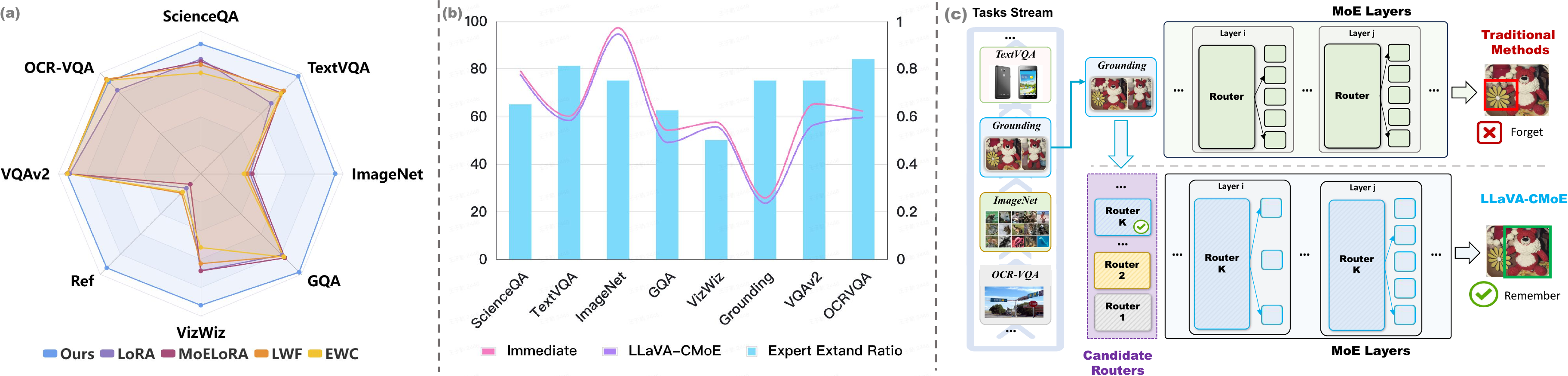}
    \caption{\textbf{a) Visualization of the Model's Anti-Forgetting Capability.} Our method significantly improves anti-forgetting capability, achieving a performance boost of approximately X\% compared to baselines. \textbf{b)  Model Forgetting Ratio vs. Parameter Expansion Rate.} \textit{Immediate} means the performance just after training. When training on task streams, our methods nearly match the performance of \textit{Immediate} while saving nearly 30\% parameters, enhancing efficiency. \textbf{c) Conceptual comparison between the previous method and our \name.} Unlike traditional methods that use a fixed number of experts, our model dynamically adjusts experts per layer based on task needs. Besides, we also build a router bank to reduce forgetting and improve knowledge retention. For clarity, we illustrate only the routing and experts within the block.}
    \label{fig:teaserv3}
\end{figure}
However, in real-world scenarios, knowledge and tasks are continuously updated in a streaming manner, posing new challenges for the efficient adaptation of MLLMs. This dynamic environment necessitates that fine-tuning methods not only remain parameter-efficient, but also support robust continual learning (CL) to enable models to incrementally acquire new knowledge from new tasks, without access to data from previous tasks. 


Catastrophic forgetting~\cite{Kirkpatrick_2017,chen2018continual}, a fundamental issue in CL, also poses a significant challenge for achieving efficient continual learning in MLLMs. To address this issue,
some approaches~\cite{cai2023task,lei2023symbolic,10.1145/3581783.3612207} involve storing data from previous tasks as subsets or distributions, and utilize a data replay scheme during new tasks training to prevent knowledge forgetting. 
However, as tasks multiply, the storage and computational overhead associated with replay-based methods can become prohibitive. 
Another line of works~\cite{10.1145/3581783.3612121,Zheng2023PreventingZT,farajtabar2020orthogonal, zhu2021counter, delchiaro2020rattrecurrentattentiontransient,2018overcomingcatastrophicforgettinghard,jha2024clap4clipcontinuallearningprobabilistic,cai2022multimodal,8994101,he2023continualinstructiontuninglarge,10.1007/s11263-020-01392-1,10.1145/3581783.3612207} focuses on designing novel loss functions or model architectures to alleviate forgetting. Nonetheless, these approaches typically involve updating parameters shared across tasks, which may still lead to performance degradation on previously learned tasks due to interference.

Recently, Wang et al.~\cite{wang2024lemoe} proposed leveraging a Mixture of Experts (MoE) architecture by introducing new experts at all layers to acquire task-specific knowledge during continual learning. While the scalability of MoE architectures makes them a promising framework for continual learning, two major challenges remain: 
1) \textbf{\textit{When and where should experts be inserted?}} 
Existing study~\cite{yang2024moralmoeaugmentedlora} shows that the number of additional parameters required varies significantly across tasks, suggesting that expert allocation should be adaptively determined by task similarity rather than statically predefined. To address this, \cite{pu2023chatgpt} investigates dynamic expert allocation by analyzing changes in expert selection distributions before and after task training. However, such distributional shifts mainly reflect changes in task preference, rather than a genuine need for new experts. As a result, current approaches may struggle to accurately determine when and where new experts are truly needed for effective continual learning.
2) \textbf{\textit{How to mitigate catastrophic forgetting in the Router?}} In MoE architectures, the Router is a crucial component responsible for dynamic expert allocation. 
While optimizing experts for new tasks may lead to forgetting previously acquired knowledge, fine-tuning the Router can also alter the routing strategies for earlier tasks, similarly resulting in catastrophic forgetting.
\cite{yu2024boosting} addresses this by fixing the number of experts and continuously adding routers to ensure the integrity of the knowledge in previous routers. 
However, as the number of tasks increases, maintaining a fixed set of experts may limit the model's capacity to accommodate tasks with substantial differences.

Building on the aforementioned challenges in expert allocation and the mitigation of catastrophic forgetting, we propose an effective framework, \name, which requires no replay data while enabling truly continual learning.
Our method facilitates adaptive parameter updates, enabling the network to autonomously determine when and where updates are necessary. These updates ensure that all previously acquired knowledge remains intact during the learning of new tasks.
To tackle the challenge in expert allocation, we propose the \textbf{Probe-Guided Knowledge Extension} algorithm, which leverages a small amount of probe data from new tasks to monitor the behavior of experts newly added at each layer.
Each expert functions as a probe; by observing the activation frequency of each expert probe, the network is able to independently determines whether additional parameters are required at each layer, avoiding unnecessary parameter growth in \cite{pu2023chatgpt}.
%
%
To tackle catastrophic forgetting, we propose the \textbf{Probabilistic Task Locator} strategy. Instead of a single shared router prone to task conflicts, each task gets a dedicated router. This design preserves prior task knowledge and routing preferences—activating a router restores its performance. Lightweight routers introduce minimal overhead.
During inference, task identity is unknown, and inputs are task-agnostic. Inspired by \cite{pu2016variational, pinheiro2021variational,An2015VariationalAB}, we use a reconstruction-based strategy: via a VAE framework, we learn task-specific distributions. The model evaluates reconstruction errors to find the best-matching task distribution, allowing PTL to infer tasks, route inputs dynamically, and maintain robust performance on unknown tasks.
We evaluate LLaVA-CMoE on the CoIN dataset \cite{chen2024coin}, a benchmark tailored to assess continual learning performance across eight distinct VQA-based tasks.
Through extensive quantitative and qualitative evaluations, along with comprehensive ablation studies, we demonstrate that our approach not only significantly mitigates catastrophic forgetting, but also promotes effective knowledge transfer and adaptation as new tasks are learned sequentially.
\label{sec:intro}
\section{Related Work}

\textbf{Continual Learning.} Continual learning aims to mitigate catastrophic forgetting~\cite{HASSABIS2017245,wu2024continuallearninglargelanguage} and enable incremental knowledge acquisition. Existing methods fall into four main categories: 1) \emph{Regularization-based} approaches~\cite{10.1145/3581783.3612121,Zheng2023PreventingZT,farajtabar2020orthogonal, zhu2021counter}, which constrain updates to important parameters; 2) \emph{Architecture-based} approaches~\cite{delchiaro2020rattrecurrentattentiontransient,2018overcomingcatastrophicforgettinghard,jha2024clap4clipcontinuallearningprobabilistic,cai2022multimodal,8994101,he2023continualinstructiontuninglarge,10.1007/s11263-020-01392-1,10.1145/3581783.3612207,yu2024boosting,chen2023lifelong}, which add task-specific components to reduce interference; 3) \emph{Replay-based} approaches~\cite{cai2023task,lei2023symbolic,10.1145/3581783.3612207,lopez2017gradient}, which store or generate samples for rehearsal; and 4) \emph{Prompt-based} methods~\cite{wang2023spromptslearningpretrainedtransformers,Qian_2023_ICCV,D_Alessandro_2023,zheng2024antiforgettingmultimodalcontinualinstruction}, which use learnable prompts to maintain performance. However, computational and storage burdens, as well as forgetting due to parameter updates, remain open challenges~\cite{chen2024coin}.

\textbf{Mixture of Experts.} The MoE architecture~\cite{riquelme2021scaling, shen2023scaling, mustafa2022multimodal} employs specialized expert networks and a gating mechanism for efficient computation. Sparsely-gated MoE~\cite{Shazeer_Mirhoseini_Maziarz_Davis_Le_Hinton_Dean_2017,lepikhin2020gshardscalinggiantmodels} has shown strong performance in LLMs, such as Mixtral 8x7B~\cite{jiang2024mixtralexperts}, across diverse NLP tasks. Recent work~\cite{liu2024moe,yang2024moralmoeaugmentedlora,chen2023octavius,Luo2024MoELoRACL} combines MoE with LoRA~\cite{hu2022lora} for more efficient training. MoE’s scalability has led to its adoption in continual learning, e.g., LEMoE~\cite{wang2024lemoe} adds experts at all layers for new tasks, and Lifelong-MoE~\cite{chen2023lifelong} trains new experts while freezing old ones. CoIN~\cite{chen2024coin} further explores MoELoRA, but existing methods still face parameter overhead and suboptimal robustness.

\textbf{Large Language Models \& PEFT.} Recently, Large Language Models~\cite{liu2023visual, touvron2023llama, chowdhery2023palm, brown2020language} have garnered widespread attention for their remarkable abilities in areas such as language generation, in-context learning, and reasoning.
To enable data- and compute-efficient adaptation for specific downstream tasks, various PEFT methods~\cite{karimi2021compacter,houlsby2019parameter,li2021prefix} have been introduced.
Among these, LoRA~\cite{hu2022lora} stands out by representing weight updates through low-rank decomposition with two matrices, keeping the original weights frozen while training only the new update matrices.
In this study, we combined LoRA with MoE for efficient continual MLLM fine-tuning.

\section{Methodology}

\begin{figure}[!t]
    \centering
    \includegraphics[width=1\linewidth]{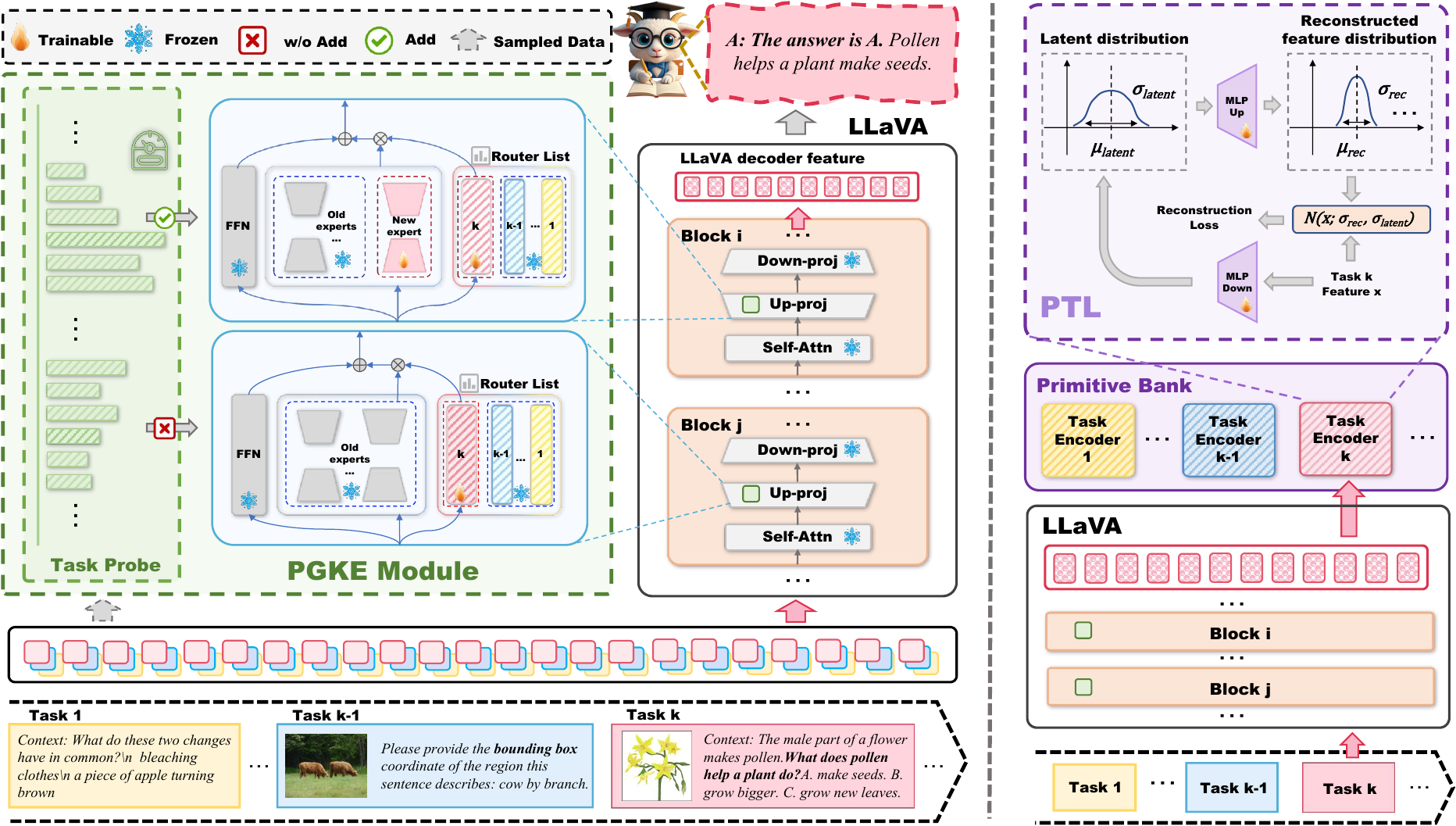}
    \caption{\textbf{Overview of our LLaVA-CMoE.} Our model consists of two main components: 1) \textbf{Probe-Guided Knowledge Expansion (PGKE)} adaptively allocates the number of experts for different tasks based on task probe guidance, enabling efficient task learning. 2) \textbf{Probabilistic Task Locator (PTL)} establishes the connection between task distributions and task routing. During inference, it identifies the corresponding router based on the input, ensuring accurate task-specific processing.}
    \label{fig:main}
\end{figure}

In this section, we first clarify the problem formulation of multimodal continual learning in Section~\ref{sec:problem_formulation}. We then elaborate on the proposed \name{} and its core components: Probe-Guided Knowledge Expansion and Probabilistic Task Locator in Section~\ref {sec:llava-cmoe}. Finally, we detail the training objectives in Section~\ref{sec:loss}.


\subsection{Problem Formulation}
\label{sec:problem_formulation}
Let $\{\mathcal{T}_1, \dots, \mathcal{T}_N\}$ be a set of $N$ tasks, where each task $\mathcal{T}_i$ has its training set $\mathbf{D}^i$ consisting of $n_i$ multimodal inputs $\mathbf{X} \in\{\mathbf{X}_{t}^i, \mathbf{X}_{img}^i, \mathbf{X}_{l}^i\}_{i=1}^{n_i}$. Here, $\mathbf{X}_{t}^i$, $\mathbf{X}_{img}^i$, and $\mathbf{X}_{l}^i$ represent the textual input (instruction), image, and the corresponding answer, respectively.
In Continual Learning, the model is trained sequentially on the $N$ tasks, and while training the $i$-th task $\mathcal{T}_i$, the model needs to maximize the probability $P_i$ through Next Token Prediction.
Notably, while training the $i$-th task, it cannot access to the data of previous tasks $\{\mathcal{T}_1, \dots, \mathcal{T}_{i-1}\}$. During the inference phase, we receive only task-agnostic data from either seen or unseen tasks. This practical constraint differentiates our approach from traditional task-incremental learning (Task-IL) paradigms, where explicit task identifiers are typically accessible during inference~\cite{shi2024continual}, thus significantly enhancing practical applicability in real-world scenarios where task boundaries may be ambiguous.
%
%
\subsection{LLaVA-CMoE} \label{sec:llava-cmoe}
In this paper, we propose \name{}, a framework that enables efficient expert expansion while maintaining strong anti-forgetting capabilities, as illustrated in Figure~\ref{fig:main}.
First, in Section~\ref{sec:moe}, we introduce the core architecture of the Mixture of Experts framework.
Next, in Section~\ref{sec:PGKE}, we detail our PGKE mechanism, which dynamically identifies optimal locations for expert expansion.
Finally, Section~\ref{sec:PTL} describes the PTL mechanism, which allows the model to effectively route inputs to task-relevant experts during inference, ensuring comprehensive knowledge utilization.
%
\subsubsection{Mixture of Expert Layer}
\label{sec:moe}
Our model is built upon a multimodal large language model (MLLM), such as LLaVA~\cite{liu2023visual}, in which the MoE is implemented by augmenting the Feed-Forward Network (FFN) modules within each Transformer block.
Furthermore, each expert is constructed using a LoRA module, parameterized by $\{\mathbf{A}_i, \mathbf{B}_i\}_{i=1}^{N_e}$, where $N_e$ denotes the number of experts.
In the $h$-th block, given the multimodal token $\mathbf{X}^{h}$, the output token $\mathbf{X}_{out}^{h}$ is computed as follows:
\begin{equation}
    \mathbf{X}_{out}^{h} = \mathbf{W}_0 \mathbf{X}^h + \sum_{i=1}^{N_e} \omega_i' \mathbf{B}_i \mathbf{A}_i \mathbf{X}^h, \quad
    \omega_i' = \frac{\omega_i}{\sum^{N_e}_{j=1}{\omega_j}}, \quad
    \omega_i = \exp(\mathbf{GX}^h)  \cdot \mathbb{I}[i \in \mathrm{topk}(\mathbf{GX}^h)],
\end{equation}
where $\mathbf{W}_0$, $\mathbf{G}$ represent the linear layers' weights of the Feed-Forward Network and the router network, respectively. $\mathbb{I}$ denotes the indicator function.
%


\subsubsection{Probe-Guided Knowledge Extension (PGKE)}
\label{sec:PGKE}

Current MoE continual learning methods usually append a {\em fixed} number of experts to every layer~\cite{wang2024lemoe,yang2024moralmoeaugmentedlora} when training a new task. 
However, simply appending a fixed number of experts to {\em every} MoE layer after each task
(i) wastes parameters when the new task is similar to previous ones,
(ii) scales quadratically with the number of tasks, and
(iii) still alters the shared router, inducing forgetting.
To address these issues, we propose
Probe-Guided Knowledge Extension (PGKE), {\bf measuring {\em where} the current model
lacks capacity {\em before} it allocates new experts}.
The key intuition is that, if the features required by the incoming task
already lie in the span of existing experts,
then a freshly initialized “probe” expert will rarely be selected.
Conversely, persistent high probe activations indicate
that {\em no existing expert can explain the new examples},
indicating that extra capacity is truly needed.
This ``try-before-you-buy’’ principle
keeps the parameter budget tight and aligns expansion with genuine
knowledge gaps.

\begin{wrapfigure}{!t}{0.65\linewidth}
    \centering
    \includegraphics[width=1\linewidth]{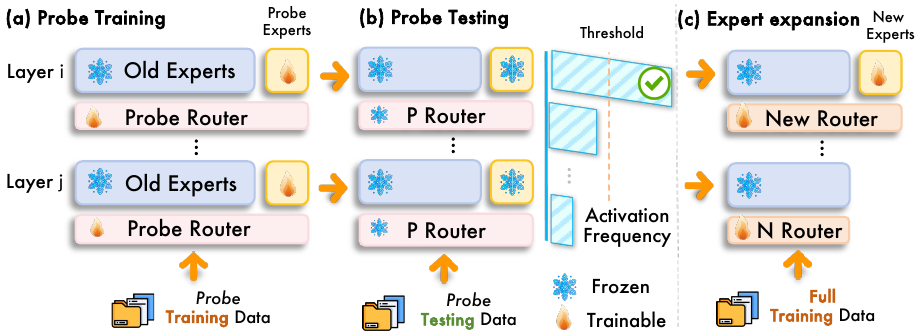}
    \caption{Illustration of Probe-Guided Knowledge Extension (PGKE) process. a) Probe experts are added to all layers. We use probe training data to train probe experts and probe routers. b) Calculating activation frequencies to select layers to expand. c) Expanding selected layers' experts and doing full training.}
    \label{fig:pkge}
    \vspace{-1mm}
\end{wrapfigure}

Following this principle, we design a two-stage training framework, consisting of \emph{probe locating} and \emph{expert expansion}. Given the training set $\mathbf{X}^i$ for the $i$-th task, we begin by sampling two non-overlapped subsets: $\mathbf{X}_\text{train}^i$ for training and $\mathbf{X}_\text{eval}^i$ for evaluation of the probe experts.

\noindent \textbf{Probe locating.}
In this phase, each MoE layer is augmented by introducing new probe experts along with a corresponding probe router, denoted as $\mathbf{R}_i$. To ensure accurate probing, $\mathbf{R}_i$ is initialized with the parameters of the router from the $(i-1)$-th task, new probe experts are initialized with average weight of old expert group. An additional dedicated sub-router is appended to accommodate the probe experts. During this phase, the parameters of all existing experts are frozen; only the parameters of the probe experts and $\mathbf{R}_i$ are updated using $\mathbf{X}_\text{train}^i$.

Upon completion of probe training, we evaluate the activation statistics of all experts in each layer over the validation set $\mathbf{X}_\text{eval}^i$ by computing both the mean and variance of their activation frequencies (logit values). Based on these statistics, we define a threshold for expert expansion as follows (with the layer index omitted for clarity):
\begin{equation}
\mathrm{Threshold}= \mathrm{mean}(\mathbf{Act}) - \alpha \cdot \mathrm{std}(\mathbf{Act}),
\end{equation}
where $\alpha$ is a hyperparameter that modulates the sensitivity of expert growth (set to 0.8). If the activation frequencies of $N_s$ probe experts exceed this threshold, we interpret this as evidence that the current layer requires additional capacity, and therefore augment the layer by introducing $N_s$ new experts accordingly. The selection process is illustrated in Figure~\ref{fig:pkge}.


%
%

\noindent \textbf{Expert expansion.}
After determining the optimal number of experts to be integrated into each layer, we perform expert expansion by augmenting the selected layers with the corresponding number of newly initialized experts and routers, following the same initialization strategy as for $\mathbf{R}_i$.
Additionally, the weights of the new expert are derived from the weights of the expert that is most frequently activated during the probe locating process.
Subsequently, the model is fine-tuned on the entire training dataset $\mathbf{X}^i$, freezing old experts to mitigate forgetting. To accommodate the varying complexities of different tasks, the parameter $N_s$, representing the number of newly added experts, is treated as a dynamically adjustable quantity.
This approach not only enhances the efficiency of parameter expansion, but also ensures that the model can effectively adapt to and learn from tasks with greater complexity.
A comprehensive analysis of its impact is also presented in Section~\ref{sec:abl}.

\subsubsection{Probabilistic Task Locator (PTL)}
\label{sec:PTL}

After continual learning, most MoE methods fall back to one
shared router that tries to satisfy all tasks.
Because the router is fine-tuned on the last task, its decision boundary therefore drifts towards the most recent data,
so earlier tasks are progressively mis-routed and forgotten.
Keeping one dedicated router per task (enabled by PGKE) would solve
this, but we must know {\em which} router to activate at test time when task labels are absent.
Training an additional classifier to predict task IDs
either (i) requires storing replay data or (ii) depends on fragile hand-crafted rules.

Drawn inspiration from \cite{pu2016variational,pinheiro2021variational,An2015VariationalAB}, we propose the {\bf Probabilistic Task Locator (PTL)}, a \emph{replay-free} mechanism that automatically selects the correct router at inference.
The hidden representation of each multimodal input is regarded as a sample from a task-specific distribution; PTL fits this distribution with a light VAE and later measures how well a test sample conforms to each task.

\noindent \textbf{Task-wise VAE fitting.}
For the $i$-th task, we pass the input through the frozen LLaVA and fetch the last-token feature $\mathbf{F}_{\text{end}}$.
The VAE encoder projects $\mathbf{F}_{\text{end}}$ to a Gaussian in the latent space as follows:
\begin{equation}\label{eq:vae_enc}
    \boldsymbol{\mu}_{\text{latent}}
        =\mathtt{FFN}_{\mu}^{\text{down}}(\mathbf{F}_{\text{end}}),\quad
    \boldsymbol{\sigma}_{\text{latent}}
        =\mathtt{Softplus}\!\bigl(
          \mathtt{FFN}_{\sigma}^{\text{down}}(\mathbf{F}_{\text{end}})\bigr),
\end{equation}
where the Softplus ensures $\boldsymbol{\sigma}_{\text{latent}}>0$.
We draw $N_{\text{rep}}$ latent codes  
$\boldsymbol{z}_{i}\sim\mathcal{N}(\boldsymbol{\mu}_{\text{latent}}, \boldsymbol{\sigma}_{\text{latent}}^{2})$
and decode them with an up-projection FFN,
obtaining the predictive parameters
$\bigl(\boldsymbol{\mu}_{\text{rec},i},\boldsymbol{\sigma}_{\text{rec},i}\bigr)$
analogous to Eq.~\eqref{eq:vae_enc}.  
The conditional likelihood of the feature can be therefore formulated as:
\begin{equation}\label{eq:vae_lik}
p(\mathbf{F}_{\text{end}}\!\mid\!\boldsymbol{z}_{i})
        =\mathcal{N}\!\bigl(
          \mathbf{F}_{\text{end}};
          \boldsymbol{\mu}_{\text{rec},i},
    \boldsymbol{\sigma}_{\text{rec},i}^{2}\bigr)
        =\frac{1}{\boldsymbol{\sigma}_{\text{rec},i}\sqrt{2\pi}}
         \exp\!\Bigl[
           -\frac{\lVert\mathbf{F}_{\text{end}}
                   -\boldsymbol{\mu}_{\text{rec},i}\rVert^{2}}{2\boldsymbol{\sigma}_{\text{rec},i}^{2}}\Bigr].
\end{equation}
Averaging Eq.~\eqref{eq:vae_lik} over $N_{\text{rep}}$ samples estimates the reconstruction probability $p_{\text{rec}}(\mathbf{F}_{\text{end}})$ for task $i$.

\noindent \textbf{Primitive bank construction.}
After the VAE convergence, we collect the most recent $T$ training instances of task $i$, compute their $p_{\text{rec}}(\mathbf{F}_{\text{end}})$, and record the empirical mean and standard deviation, representing the primitive distribution of task $i$. Consequently, we construct a key–value bank
$B = \left\{ \left(\text{primitive}_i,\, \text{router}_i\right) \mid i = 1, \ldots, N \right\}$,
where each key stores the probability statistics and each value is the frozen router produced by PGKE.

\noindent \textbf{Task-agnostic inference.}
For an input, we obtain its feature $\mathbf{F}_{\text{end}}$, evaluate and z-score normalise its reconstruction probability under every primitive in $B$, yielding scores $\{\hat p^{(i)}\}_{i=1}^{N}$.
The task is inferred as  
$
    i^{\star}=\arg\max_{i}\hat p^{(i)},
$
and the corresponding $\text{router}_{(i^{\star})}$ is activated for subsequent processing.

\subsection{Training Objective}
\label{sec:loss}

For PTL, the training loss consists of two components: the reconstruction loss and the Kullback-Leibler (KL) divergence between the posterior distribution and the prior distribution of the latent space. The reconstruction loss can be measured by the negative reconstruction probability:
\begin{equation}
    \mathcal{L}_{\text{rec}}=-{p}_{\text{rec}}(\mathbf{F}_\text{end}).
\end{equation}

We follow the common setting of VAE that the prior distribution of the latent space is $ \mathcal{N}(\boldsymbol{0},\boldsymbol{1})$\cite{Kingma2013AutoEncodingVB}. The posterior distribution of the latent space is $\mathcal{N}(\boldsymbol{\mu}_\mathrm{latent}, \boldsymbol{\sigma}_\mathrm{latent}^2)$. The KL divergence $\mathcal{L}_{\text{KL}}$ is calculated between the two distributions.
For PGKE, similar to most large language models, we employ the next token prediction training schema and utilize a cross-entropy loss $\mathcal{L}_{\text{CE}}$.
Besides, referring to \cite{fedus2022switchtransformersscalingtrillion}, we also add a weight balance loss $\mathcal{L}_{aux}$ for MoE training.
Finally, the complete loss is composed of a weighted sum of these components:
\begin{equation}
    \mathcal{L}_{\text{total}} = \mathcal{L}_{\text{CE}} + \lambda  \mathcal{L}_{\text{KL}} + \eta  \mathcal{L}_{\text{rec}} + \kappa  \mathcal{L}_{\text{aux}},
\end{equation}
where $\lambda$, $\eta$ and $\kappa$ are the weighting coefficients for loss balance. 
\section{Experiments and Discussions}
\subsection{Setups and Implementation Details.}
\begin{table}[!t]
\caption{A comprehensive comparison with baseline models and other continual learning approaches built upon LLaVA is detailed in the subsequent section. \textit{Immediate} means performance after immediate task training. \textit{Last} means performance after the last task training. \textit{Mean} means the average accuracy on all eight tasks.}
\label{exp:main}
\renewcommand\arraystretch{1.1}
\renewcommand\tabcolsep{2.0pt}
\centering
\resizebox{\linewidth}{!}{
\begin{tabular}{llcccccccccc}
\toprule
\multirow{2}{*}{\textbf{Setting}} &
\multirow{2}{*}{\textbf{Method}} &
\multicolumn{8}{c}{\textbf{Accuracy on Each Task}} &
\multicolumn{1}{c}{\multirow{2}{*}{Mean$ \uparrow $}} &
\multicolumn{1}{c}{\multirow{2}{*}{BWT$ \uparrow $}} \\ \cmidrule(r){3-10} 
& & SQA & TQA & ImageNet & GQA & VizWiz & Ref & VQAv2 & OCR-VQA &  &  \\ 
\midrule

\multirow{1}{*}{Multitask} & {MoELoRA} & 75.01 & 58.90 & 96.44 & 58.15 & 56.73 & 27.54 & 64.04 & 47.81 & 60.58 & --  \\
\midrule

\multirow{4}{*}{Immediate} & {LoRA} & 75.01 & 58.36 & 96.08 & 53.87 & 56.54 & 18.32 & 63.23 & 53.96 & 59.42 & --  \\
& {MoELoRA} & 78.97 & 61.01 & 97.01 & 56.24 & 56.30 & 20.63 & 66.10 & 60.57 & 62.10 & --  \\
& {EWC~\cite{schwarz2018progress}} & 79.23 & 61.26 & 96.91 & 56.43 & 60.04 & 19.21 & 66.00 & 60.44 & 62.44 & -- \\
& {LWF~\cite{li2017learning}} & 78.83 & 61.57 & 97.07 & 56.75 & 53.48 & 20.57 & 65.27 & 61.10 & 61.83 & -- \\
& {Ours} & \textbf{79.01} & \textbf{59.94} & \textbf{96.85} & \textbf{56.43} & \textbf{57.44} & \textbf{25.63} & \textbf{65.15} & \textbf{62.01} & \textbf{62.81} & -- \\ 
\midrule

\multirow{4}{*}{Last} & {LoRA} & 68.50 & 42.01 & 34.65 & 40.39 & 40.87 & 3.60 & 55.29 & 53.96 & 42.41 & -17.01  \\
& {MoELoRA} & 67.06 & 48.16 & 36.22 & 41.83 & 41.00 & 2.62 & \textbf{56.48} & 60.57 & 44.24 & -17.86\\

& {EWC~\cite{schwarz2018progress}} & 60.25 & 47.92 & 30.36 & 41.33 & 31.12 & 4.53 & 56.31 & 60.44 & 41.53 & -20.90 \\
& {LWF~\cite{li2017learning}} & 65.15 & 49.46 & 32.52 & 41.05 & 37.88 & 4.81 & 56.20 & \textbf{61.10} & 43.51 & -18.31 \\
& {Ours} & \textbf{77.55} & \textbf{58.17} & \textbf{94.50} & \textbf{48.91} & \textbf{55.45} & \textbf{23.40} & 56.40 & 59.44 & \textbf{59.23} & \textbf{-3.58} \\ 
\bottomrule




\end{tabular}
}
\vspace{3mm}
\end{table}
\textbf{Datasets.}
We conducted experiments on the datasets included in the CoIN~\cite{chen2024coin} benchmark, which encompasses a series of eight VQA tasks. These tasks include RefCOCO (Ref)~\cite{kazemzadeh-etal-2014-referitgame}, ImageNet~\cite{5206848}, TextVQA (TQA)~\cite{DBLP:conf/cvpr/SinghNSJCBPR19}, VizWiz~\cite{8578478}, ScieneQA (SQA)~\cite{saikh2022scienceqa}, among others. Each task varies in terms of the number of data samples, stylistic features, and domain characteristics. The training set comprises a total of 569k samples, while the testing set contains 261k samples.

\textbf{Metrics.}
We adopt the metric introduced in CoIN~\cite{chen2024coin}, which measures the discrepancy between the model's output and the ground truth.
For assessing the model's overall forgetting performance across all tasks, we utilized Backward Transfer (BWT), which evaluates the model's performance on all previous tasks after it is trained on the last task, specifically quantifies the extent of forgetting, offering insights into the model's ability to retain knowledge from previous tasks.

\textbf{Baseline Models.} Following CoIN~\cite{chen2024coin}, we also adopted several other representative methods based on architecture and regularization, including EWC~\cite{schwarz2018progress}, LwF~\cite{li2017learning}, as baselines to compare with our proposed method. 
To ensure a fair comparison, we align factors that could potentially affect fairness, such as the rank of LoRA and the data used for initializing the model.
For more details, refer to CoIN.

\textbf{Training Details.}
Our network is built upon the pretrained model of LLaVA as our backbone model. 
It is important to emphasize that throughout all our training processes, only the newly added experts and the corresponding routers are trained, while other existing components remain frozen. 
New expert weights are copied from the experts selected by Task Probe, and new router weights are initialized by copying old routers and concatenating them with random linear weights.
While training Task-Probe, we randomly select 10\% data from each train dataset and utilize them to generate probability of each layer to extend.
Please refer to the Appendix for details of the network structures, hyperparameters, and any other implementation details.
%

\subsection{Quantitative Results on Continual Learning Benchmark}
As presented in Table~\ref{exp:main}, we conducted an evaluation on the CoIN~\cite{chen2024coin} Benchmark. 
%
%
Compared with other methods, our model demonstrates outstanding performance in both immediate and post-last-task training phases, while the average number of trainable parameters in our model is only 43.96M, more details can be found in the appendix. This significantly reduces the training cost.
In the setting of \textit{Last}, our method significantly outperforms previous approaches, demonstrating its strong capability in mitigating forgetting, especially on the ImageNet dataset, our method achieves an improvement of nearly 85\%.
Additionally, it is observed that on the OCR-VQA dataset, all other methods yield consistent results across both settings. 
However, when a new task is introduced, these methods exhibit substantial forgetting on this dataset, as evidenced by the performance in the first seven tasks. 
In contrast, our method maintains a stable performance of approximately 59\%, showcasing its robustness against forgetting.
Besides, as evident from the left of Table \ref{tab:abl_1}, employing our proposed method for expert extension and training yields superior results on most tasks compared to the outcomes of training without expanding experts and unfreezing the previously frozen experts.
%

%


\begin{table}[!t]
\centering
\caption{\textbf{Left}: Comparison of forgetting between the Extend and w/o Extend models. “Extend” freezes original experts and adds new ones using our method, while “w/o Extend” unfreezes all experts and continues training without adding new ones.
\textbf{Right}: Comparison of expert addition strategies. “Ours” adds experts at selected layers using our method, “Random” adds experts randomly at the same number of layers, and “Every-layer” adds experts to all layers. Param-ratio is the proportion of parameters added by Ours compared to Every-layer.}
\label{tab:abl_1}
\vspace{5pt}
\begin{subtable}[t]{0.383\linewidth}
\centering
\renewcommand\arraystretch{1.15}
\renewcommand\tabcolsep{6.0pt}
\resizebox{\linewidth}{!}{
\begin{tabular}{lcc}
    \toprule
    \textbf{Dataset} & \emph{w/o} \textbf{Extend} & \textbf{Extend} \\
    \midrule
    SQA & 76.68 & \textbf{77.55} \\
    TQA & 49.42 & \textbf{58.17} \\
    ImageNet & 45.72 & \textbf{94.50} \\
    GQA & 44.65 & \textbf{48.91} \\
    VizWiz & 46.79 & \textbf{55.45} \\
    Ref & 4.00 & \textbf{23.40} \\
    VQAv2 & \textbf{58.58} & 56.40 \\
    OCR-VQA & \textbf{60.24} & 59.44 \\
    \midrule
    BWT$\uparrow$ & -17.68 & \textbf{-3.58} \\
    \bottomrule

\end{tabular}}
\end{subtable}
\hfill
\begin{subtable}[t]{0.59\linewidth}
    \centering
    \renewcommand\arraystretch{1.3}
    \renewcommand\tabcolsep{4.0pt}
    \resizebox{\linewidth}{!}{
    \begin{tabular}{lcccc}
        \toprule
        \textbf{Dataset} & \textbf{Random} & \textbf{Every-layer} & \textbf{Ours} & \textbf{Param-ratio} \\
        \midrule
        {SQA} & 76.73 & \textbf{80.03} & 79.01 & 0.65\\
        {TQA} & 58.87 & \textbf{60.07} & 59.94 & 0.812\\
        {ImageNet} & 96.75 & \textbf{97.09} & 96.85 &0.75\\
        {GQA} & \textbf{57.60} & 57.28 & 56.43 & 0.625\\
        {VizWiz} & 54.62 & 56.52 & \textbf{57.44} & 0.5\\
        {Ref} & 20.98 & \textbf{31.68} & 25.63 & 0.75\\
        {VQAv2} & 64.20 & 64.97 & \textbf{65.15} & 0.75\\
        {OCR-VQA} & 56.87 & 59.78 & \textbf{62.01} & 0.84\\
        \bottomrule
    \end{tabular}}
\end{subtable}
\vspace{-6pt}
\end{table}
\subsection{Ablation Study and Analysis}
\label{sec:abl}
\textbf{\textit{Where should the experts be extended?}}
When expanding experts, a common practice is to add experts to every layer for each task~\cite{yang2024moralmoeaugmentedlora}. 
However, we argue that this approach is inefficient due to significant knowledge overlap between tasks, which leads to parameter redundancy. 
To validate the effectiveness of our PGKE algorithm, as shown in Table~\ref{tab:abl_2}, we compare several expansion strategies.
By comparing the results of Ours and Every-layer, it is evident that our method achieves comparable performance with only 60\% of the parameters, even outperforming the every-layer approach on OCR-VQA. 
Furthermore, the comparison between Random and Ours demonstrates that the positions identified by the task probe are more reasonable, yielding an average performance improvement of 3\%-4\% under the same training parameter budget, showing the superiority of the PGKE method.
%

\begin{table}[!t]
\vspace{-4pt}
\caption{Comparison of different task orders. ``Norm'' denotes that we follow CoIN's training order. ``Rev'' denotes that we reverse CoIN's training order. ``Rand'' denotes that we random shuffle the training order of the CoIN's datasets.}
\label{tab:abl_2}
    \centering
    \renewcommand\arraystretch{1.1}
    \renewcommand\tabcolsep{8.0pt}
    \resizebox{0.7\linewidth}{!}{
    \begin{tabular}{lcccccc}
        \toprule
        \multirow{2}{*}{\textbf{Setting}} &
        \multicolumn{3}{c}{\textbf{Immediate}} &
        \multicolumn{3}{c}{\textbf{Last}} \\ 
        \cmidrule(r){2-4} \cmidrule(r){5-7}
        & Norm & Rev & Rand & Norm & Rev & Rand \\
        \midrule
        ScienceQA & 79.01 & \textbf{79.96} & 79.63 & 77.55 & 78.42 & \textbf{78.48}\\
        TextVQA & \textbf{59.94} & 58.60 & 58.51 & \textbf{58.17} & 56.12 & 56.01\\
        ImageNet & 96.85 & \textbf{96.99} & 96.89 & 94.50 & \textbf{94.61} & 94.55\\
        GQA & 56.43 & \textbf{57.88} & 57.81 & 48.91 & \textbf{50.32} & 50.29\\
        VizWiz & 57.44 & \textbf{58.76} & 57.92 & 55.45 & \textbf{56.77} & 56.70\\
        Ref & 25.63 & \textbf{29.29} & 29.13 & 23.40 & \textbf{27.23} & 27.09\\
        VQAv2 & \textbf{65.15} & 64.75 & 64.83 & \textbf{56.40} & 56.02 & 56.16\\
        OCR-VQA & \textbf{62.01} & 61.27 & 61.31 & \textbf{59.44} & 58.72 & 58.66\\
        \midrule
        Mean $\uparrow$ & 62.81 & \textbf{63.43} & 63.25 & 59.23 & \textbf{59.78} & 59.74\\ 
        BWT $\uparrow$ & - & - & - & -3.58 & -3.66 & \textbf{-3.51} \\
        \bottomrule
    \end{tabular}}
\vspace{4pt}
\end{table}

\textbf{\textit{How does task order influence performance?}} 
We conducted experiments on our method under different training orders across eight datasets of CoIN to verify its robustness to training order. 
As shown in the table, the forgetting level of our method remains stable across different training orders. 
This is because the feature extraction and reconstruction of different tasks are independent, same as task routers.
Previously trained tasks do not affect the features of subsequent tasks. 
Additionally, training order has a minimal impact on the method's immediate performance, though this is not our primary focus. More details will be discussed in the Appendix.

\textbf{\textit{How many experts should be extended?}}
Our experiments show that for certain challenging tasks, increasing the number of parameters is crucial, even when added to layers with overlapping knowledge (see the left side of Table~\ref{tab:abl_1}). To investigate further, we examined how the number of experts per layer affects performance on the Ref and SQA tasks by adding 1, 4, or 8 experts to probe-selected layers (results in Table~\ref{tab:abl_3}). We found that performance on the more difficult Grounding task improves significantly with more parameters, while the simpler SQA task benefits only marginally. However, due to the distinction between task difficulty and task dissimilarity, we could not develop an end-to-end method for determining the optimal number of experts based on task difficulty. As a result, we focused on optimizing layer selection and treated the number of experts as a hyperparameter.

\begin{wraptable}{R}{6.5cm}
\vspace{-2pt}
\caption{The impact of the number of experts added at specified layers on the final performance.}
\label{tab:abl_3}
    \centering
    \renewcommand\arraystretch{1.25}
    \renewcommand\tabcolsep{4.0pt}
    \resizebox{1\linewidth}{!}{
    \begin{tabular}{lccc}
        \toprule
        \textbf{Number of Added Experts} & \textbf{1} & \textbf{4} & \textbf{8} \\
        \midrule
        {Ref} & 25.63 & 35.13  & 41.51\\
        {ScienceQA} & 79.01 & 81.42  & 82.01\\
        \bottomrule
    \end{tabular}}
\end{wraptable}

\begin{figure}[ht]
    \raggedright
    \begin{minipage}[t]{0.3\textwidth}
        \centering
        \caption{Qualitative Results of LLaVA-CMoE.}
        \label{fig:vis_2}        \includegraphics[width=1\linewidth]{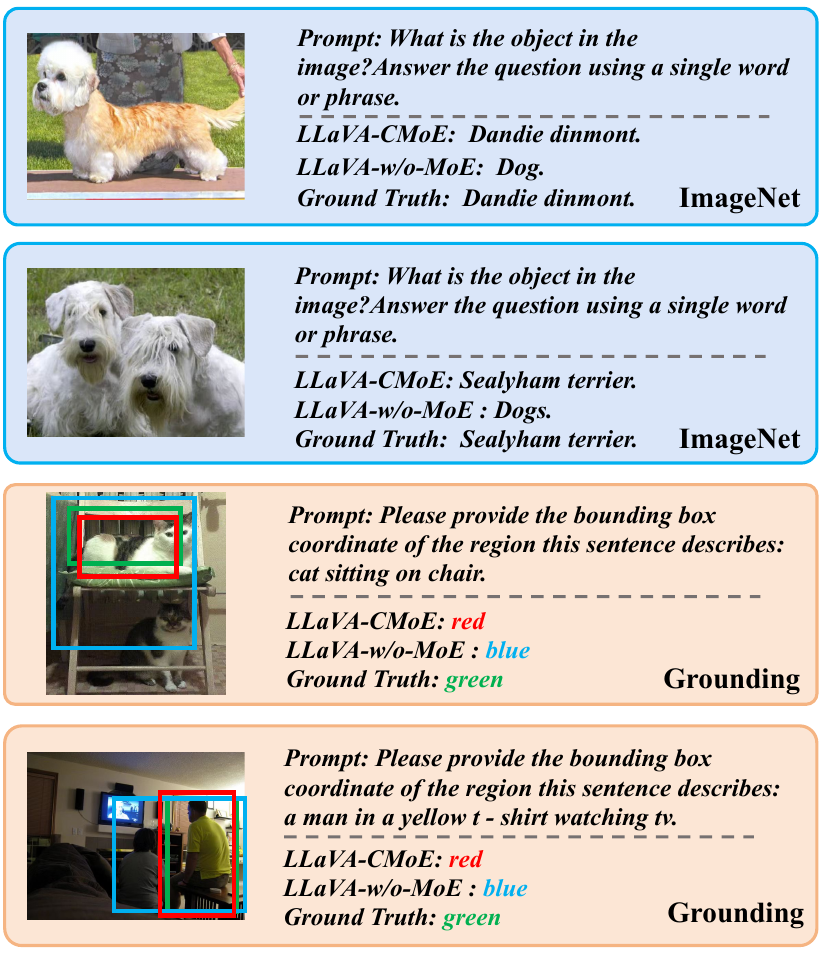}
        \captionsetup{type=figure}
    \end{minipage}%
    \hspace{0.02\textwidth}
    \begin{minipage}[t]{0.675\textwidth}
        \centering
        \captionsetup{type=table}
        \caption{Left: Comparison of different task classification strategies. Right: Knowledge forward transfer ability comparison.}
        \label{tab:abl_4}
        \vspace{0pt} 
        \begin{subtable}[t]{0.51\linewidth}
            \centering
            \renewcommand\arraystretch{1.25}
            \renewcommand\tabcolsep{2.0pt}
            \resizebox{1\linewidth}{!}{
            \begin{tabular}{lccc}
                \toprule
                \textbf{Dataset} & \textbf{Last} & \textbf{Random} & \textbf{Ours} \\
                \midrule
                SQA & 73.03 & 61.92 & \textbf{77.55}\\
                TQA & 46.82 & 50.62 & \textbf{58.17}\\
                ImageNet & 29.68 & 36.59 & \textbf{94.50}\\
                GQA & 41.81 & 43.12 & \textbf{48.91}\\
                VizWiz & 44.32 & 37.86 & \textbf{55.45}\\
                Ref & 9.92 & 4.83 & \textbf{23.40}\\
                VQAv2 & 55.13 & 45.90 & \textbf{56.40}\\
                OCR & 62.01 & 25.62 & \textbf{59.44}\\
                \bottomrule
            \end{tabular}}
        \end{subtable}
        \begin{subtable}[t]{0.46\linewidth}
            \centering
            \renewcommand\arraystretch{1.25}
            \renewcommand\tabcolsep{4.0pt}
            \resizebox{1\linewidth}{!}{
            \begin{tabular}{lcc}
                \toprule
                \textbf{Dataset} & \textbf{Separate} & \textbf{Ours} \\
                \midrule
                SQA & 78.97 & \textbf{79.01} \\
                TQA & \textbf{60.56} & 59.94 \\
                ImageNet & \textbf{97.05} & 96.85 \\
                GQA & 56.31 & \textbf{56.43} \\
                VizWiz & 56.20 & \textbf{57.44} \\
                Ref & 21.3 & \textbf{25.63} \\
                VQAv2 & 65.01 & \textbf{65.15} \\
                OCR & 60.79 & \textbf{62.01} \\
                \bottomrule
            \end{tabular}}
        \end{subtable}
    \end{minipage}
    \hspace{-1cm}
\vspace{-3mm}
\end{figure}

\textbf{\textit{How does the PTL mechanism perform?}}
To figure out the PTL mechanism, we conducted the following two ablation experiments:
1) \textbf{Last} task evaluation: We utilized the router and the corresponding set of experts learned from the last task to evaluate all previous tasks.
2) \textbf{Random} task evaluation: We randomly selected a task router and its corresponding set of experts to evaluate all tasks.
Results are presented in the left part of Table~\ref{tab:abl_4}. The last task is excluded from analysis since it is not affected by forgetting. Under these conditions, our PTL method significantly outperforms the other approaches, indicating that our task reconstruction strategy enables PTL to better retain knowledge of past tasks. Additional experiments on feature extraction methods are detailed in the Appendix.

\textbf{\textit{Can knowledge from previous tasks facilitate new task learning?}} In continual learning, beyond mitigating forgetting, it is also important to assess whether prior knowledge facilitates learning new tasks. To this end, we evaluated the model’s forward transfer by comparing it to models trained from scratch on each of the eight tasks, with an equal number of trainable parameters, as shown in the right side of Table~\ref{tab:abl_4}. Results show that while prior knowledge offers limited benefits for simpler tasks, it significantly accelerates learning on more challenging tasks like Ref and OCRVQA.

\paragraph{Qualitative Results.} 

We qualitatively analyze the model’s outputs. As illustrated in Figure~\ref{fig:vis_2}, after training on the final task, we randomly sample data from previous tasks and compare results across methods. On ImageNet, our model retains domain-specific knowledge with minimal forgetting, while LLaVA-w/o-MoE relies on pretrained knowledge and produces generic responses. For the Grounding task, our model better preserves knowledge required for non-linguistic generation.

\section{Conclusion and Discussion}

In this paper, we propose LLaVA-CMoE, a continual learning framework comprising two key modules: Probe-Guided Knowledge Extension (PGKE) and  Probabilistic Task Locator (PTL). PGKE addresses the inefficiency of continuous parameter expansion by adaptively increasing parameters through probe-guided expert addition. Meanwhile, PTL mitigates catastrophic forgetting in continual learning by modeling task distributions and memorizing the mapping between task distributions and router networks.
Qualitative and quantitative results demonstrate that our method remarkably outperforms the existing methods.

\noindent \textbf{Limitation}. Expert addition is currently limited, as adding experts only to the language model may not fully benefit tasks requiring detailed visual understanding. Additionally, increasing tasks raises storage demands, and existing distillation or merging methods can cause router-related forgetting. We will address these issues in future work.

\noindent \textbf{Societal impact}. Our method improves continual learning in multimodal models, benefiting applications like education and accessibility. However, increased adaptability may also heighten risks of misuse, underscoring the need for responsible deployment and oversight.

\newpage
\bibliographystyle{plain}
\bibliography{main}

\newpage
\appendix
\section{Technical Appendices and Supplementary Material}

\subsection{More Results}

\begin{figure}[h]
    \centering
    \includegraphics[width=0.7\linewidth]{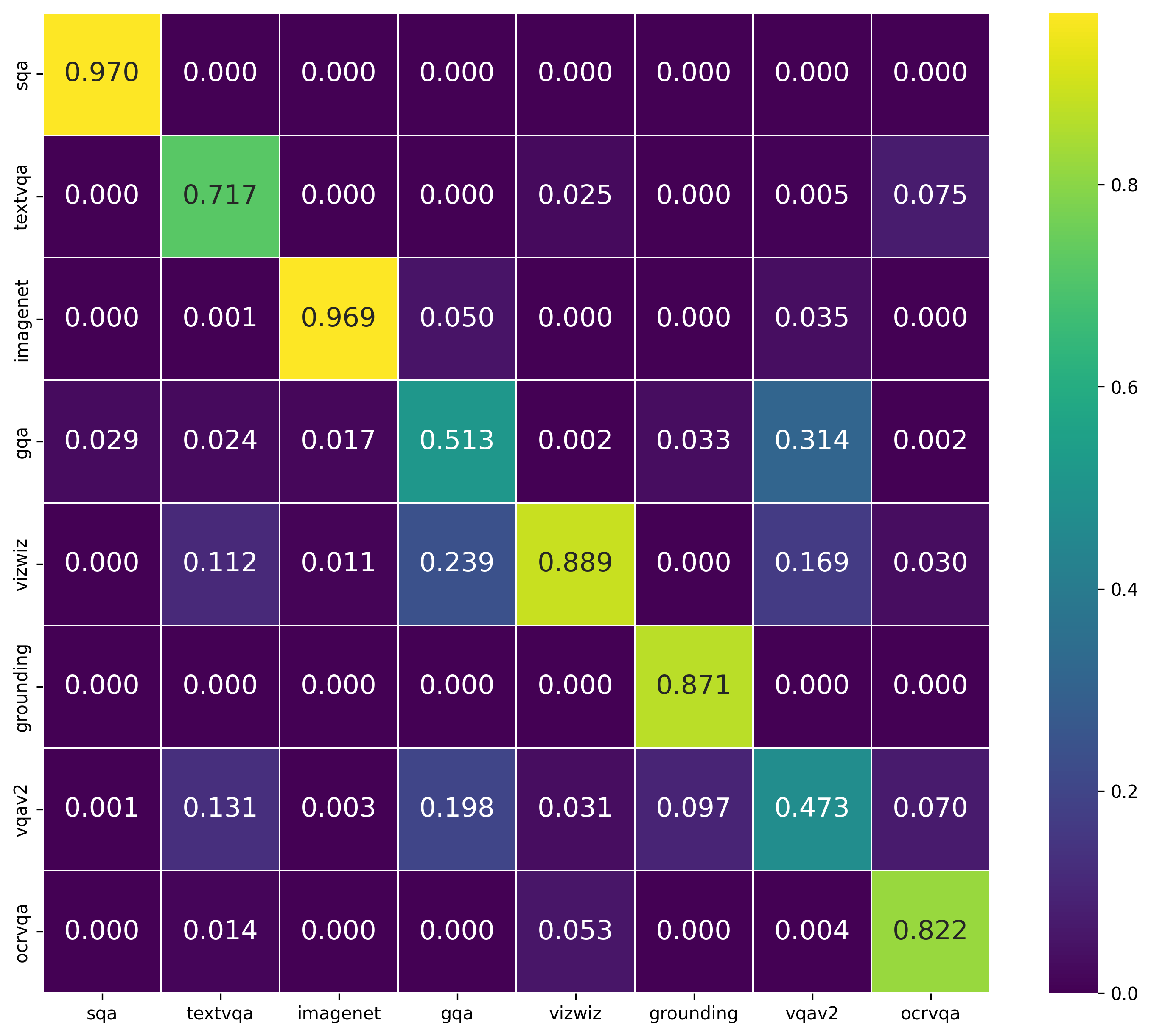}
    \caption{\textbf{Confusion Matrix of PTL.} This figure illustrates the localization performance of the PTL mechanism, where the data in each row and column represent the frequency with which the test data corresponding to the task represented by the column is assigned to the task represented by the row.}
    \label{fig:ptl}
\end{figure}

\paragraph{Visualization of PTL mechanism.}
We evaluated the PTL mechanism on test data across eight tasks, and the results are shown in Figure~\ref{fig:ptl}. As observed from the test results, the PTL mechanism achieves localization accuracies exceeding 80\% on five tasks: ScienceQA, ImageNet, VizWiz, Grounding and OCR-VQA. Besides, we can also find that the localization performance on the remaining three tasks, GQA, TextVQA, and VQAv2, is relatively weaker compared to the other five tasks. Through our analysis, this is attributed to the significant overlap in the features of images and questions across these tasks. For instance, GQA and VQAv2 share high similarities in terms of question formats, image content, and styles, which leads to a substantial portion of VQAv2 samples being ``mislocalized" to the GQA task.

\paragraph{Impact of different selection of PTL feature.}
In our main results, we chose the output of the last Decoder layer for classification in the PTL. In Table~\ref{sup:tab_1}, we also attempted to use the average pooling of outputs from all Decoder layers for classification. Classifying using features from average pooling slightly improved performance on several tasks (SQA, TQA, GQA, VizWiz, Ref), particularly achieving a 7.03\% improvement on VQAv2. However, severe performance degradation ($\sim$39\%) occurred when facing OCR-VQA. We further explored the causes of this decline and found that most samples were classified into the ImageNet task, while a small number of ImageNet samples were classified into OCR-VQA, indicating that average pooling is poor in distinguishing highly similar tasks.

\begin{table}[h]
\caption{Comparison of different feature selection’s influence on model's performance.}
\label{sup:tab_1}
\renewcommand\arraystretch{1.1}
\renewcommand\tabcolsep{2.0pt}
\centering
\resizebox{\linewidth}{!}{
\begin{tabular}{lcccccccccc}
\toprule
\multirow{2}{*}{\textbf{Method}} &
\multicolumn{8}{c}{\textbf{Accuracy on Each Task}} &
\multicolumn{1}{c}{\multirow{2}{*}{Mean$ \uparrow $}} &
\multicolumn{1}{c}{\multirow{2}{*}{BWT$ \uparrow $}} \\ \cmidrule(r){2-9} 
& SQA & TQA & ImageNet & GQA & VizWiz & Ref & VQAv2 & OCR-VQA &  &  \\ 
\midrule
{Last Layer's feature} & 77.55 & 58.17 & 94.50 & 48.91 & 55.45 & 23.40 & 56.40 & 59.44 & 59.23 & -3.58  \\
{Pooled feature} & 77.67 & 59.19 & 89.03 & 51.06 & 57.03 & 24.69 & 63.43 & 20.68 & 55.34 & -15.31 \\
\bottomrule
\end{tabular}
}
\vspace{3mm}
\end{table}

\paragraph{Impact of model size.}
We conducted experiments on models of varying scales, as delineated in Table~\ref{sup:tab_2}. Owing to the augmentation in the parameter count of the base model, the performance of the proposed method was further enhanced. Notably, the BWT metric of our approach exhibits a higher value on the 13B model. This phenomenon may be attributed to the fact that the output of the final decoder layer in larger models encompasses more highly abstracted information (hidden-size: 5120 in the 13B model vs. 4096 in the 7B model), which facilitates PTL in achieving superior task classification performance.
\begin{table}[h]
\caption{Comparison of our method on llava-v1.5-7B and llava-v1.5-13B.}
\label{sup:tab_2}
\renewcommand\arraystretch{1.1}
\renewcommand\tabcolsep{2.0pt}
\centering
\resizebox{\linewidth}{!}{
\begin{tabular}{lccccccccccc}
\toprule
\multirow{2}{*}{\textbf{Setting}} &
\multirow{2}{*}{\textbf{Model Size}} &
\multicolumn{8}{c}{\textbf{Accuracy on Each Task}} &
\multicolumn{1}{c}{\multirow{2}{*}{Mean$ \uparrow $}} &
\multicolumn{1}{c}{\multirow{2}{*}{BWT$ \uparrow $}} \\ \cmidrule(r){3-10} 
& & SQA & TQA & ImageNet & GQA & VizWiz & Ref & VQAv2 & OCR-VQA &  &  \\ 
\midrule

\multirow{2}{*}{Immediate} & {7B} & 79.01 & 59.94 & 96.85 & 56.43 & 57.44 & 25.63 & 65.15 & 62.01 & 62.81 & --  \\
& {13B} & 82.32 & 65.77 & 98.26 & 60.41 & 62.35 & 30.91 & 68.75 & 67.92 & 67.08 & --  \\
\midrule

\multirow{2}{*}{Last} & {7B} & 77.55 & 58.17 & 94.50 & 48.91 & 55.45 & 23.40 & 56.40 & 59.44 & 59.23 & -3.58  \\
& {13B} & 80.79 & 64.82 & 96.01 & 52.93 & 60.86 & 29.11 & 62.18 & 62.27 & 63.62 & -3.46\\

\bottomrule
\end{tabular}
}
\vspace{3mm}
\end{table}

\paragraph{Impact of LoRA-rank.}

We further investigated the impact of the LoRA rank on our method, as presented in the Table~\ref{sup:tab_3}. When the LoRA rank was set to 32, there remained a certain margin for performance improvement in the model. However, when the LoRA rank reached 128, the model performance had reached a plateau, with minor performance degradation observed on specific tasks. Therefore, we selected a LoRA rank of 64 as the default setting in our method.

\begin{table}[h]
\caption{Comparison of different lora rank.}
\label{sup:tab_3}
\renewcommand\arraystretch{1.1}
\renewcommand\tabcolsep{2.0pt}
\centering
\resizebox{\linewidth}{!}{
\begin{tabular}{lccccccccccc}
\toprule
\multirow{2}{*}{\textbf{Setting}} &
\multirow{2}{*}{\textbf{Rank}} &
\multicolumn{8}{c}{\textbf{Accuracy on Each Task}} &
\multicolumn{1}{c}{\multirow{2}{*}{Mean$ \uparrow $}} &
\multicolumn{1}{c}{\multirow{2}{*}{BWT$ \uparrow $}} \\ \cmidrule(r){3-10} 
& & SQA & TQA & ImageNet & GQA & VizWiz & Ref & VQAv2 & OCR-VQA &  &  \\ 
\midrule

\multirow{3}{*}{Immediate} & {32} & 78.86 & 59.90 & 96.32 & 55.92 & 57.06 & 24.17 & 63.15 & 60.22 & 61.95 & --  \\
& {64} & 79.01 & 59.94 & 96.85 & 56.43 & 57.44 & 25.63 & 65.15 & 62.01 & 62.81 & --  \\
& {128} & 79.23 & 59.54 & 97.07 & 57.96 & 56.03 & 25.58 & 65.22 & 62.33 & 62.75 & --  \\
\midrule

\multirow{3}{*}{Last} & {32} & 77.32 & 58.11 & 94.06 & 48.44 & 55.07 & 22.08 & 54.02 & 57.80 & 58.36 & -3.59  \\
& {64} & 77.55 & 58.17 & 94.50 & 48.91 & 55.45 & 23.40 & 56.40 & 59.44 & 59.23 & -3.58  \\
& {128} & 77.70 & 57.76 & 94.59 & 50.06 & 54.12 & 23.11 & 56.49 & 59.66 & 59.19 & -3.56\\
\bottomrule
\end{tabular}
}
\vspace{3mm}
\end{table}

\begin{figure}[!t]
    \centering
    \includegraphics[width=0.8\linewidth]{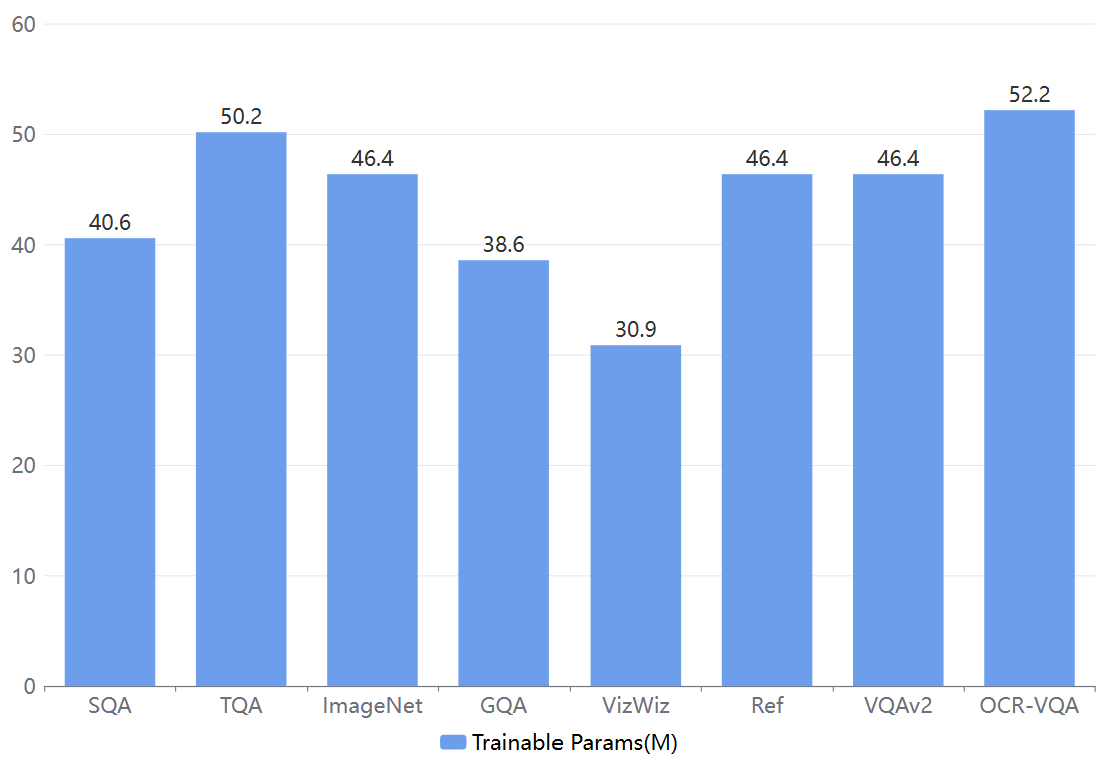}
    \caption{The added trainable parameters on eight tasks.}
    \label{fig:params}
\end{figure}

\paragraph{Parameter comparsion.}
We tabulated the number of parameters for different methods, as shown in Table~\ref{sup:tab_4}. It is worth noting that the LoRA method we compared only adds LoRA matrices to the up-proj layer. Although the LoRA method has the fewest trainable parameters, it exhibits poor anti-forgetting performance. Our method adaptively adds experts based on task differences, significantly enhancing anti-forgetting capabilities while maintaining minimal parameter growth. The changes in trainable parameters across different tasks are illustrated in Figure~\ref{fig:params}.

\begin{table}[!t]
\centering
\caption{Comparison of the average trainable parameters (M) over eight datasets, and performance between LoRA, MoELoRA, EWC, LWF and Ours.}
\label{sup:tab_4}
\renewcommand\arraystretch{1.1}
\renewcommand\tabcolsep{6.0pt}
\resizebox{0.8\linewidth}{!}{
\begin{tabular}{lccccc}
    \toprule
    \textbf{Metrics} & LoRA & MoELoRA & EWC & LWF & Ours \\
    \midrule
    {\#Trainable Parameters per Task}  & 31M & 62M & 62M  & 62M & 43.96M \\ 
    {Mean$ \uparrow $} & 42.41 & 44.24 & 41.53 & 43.51 & 59.23 \\
    {BWT$ \uparrow $} & -17.01 & -17.86 & -20.90 & -18.31 & -3.58 \\ 
    \bottomrule
\end{tabular}}
\end{table}

\paragraph{Qualitative Results.} 
We qualitatively analyze the model’s outputs. As illustrated in Figure~\ref{fig:vis_2}, after training on the final task, we randomly sample data from previous tasks and compare results across methods. On ImageNet, our model retains domain-specific knowledge with minimal forgetting, while MoELoRA and other methods rely on pretrained knowledge and produces generic responses. For the Ref task, our model better preserves knowledge required for non-linguistic generation. 

\begin{figure}[!t]
    \centering
    \includegraphics[width=1\linewidth]{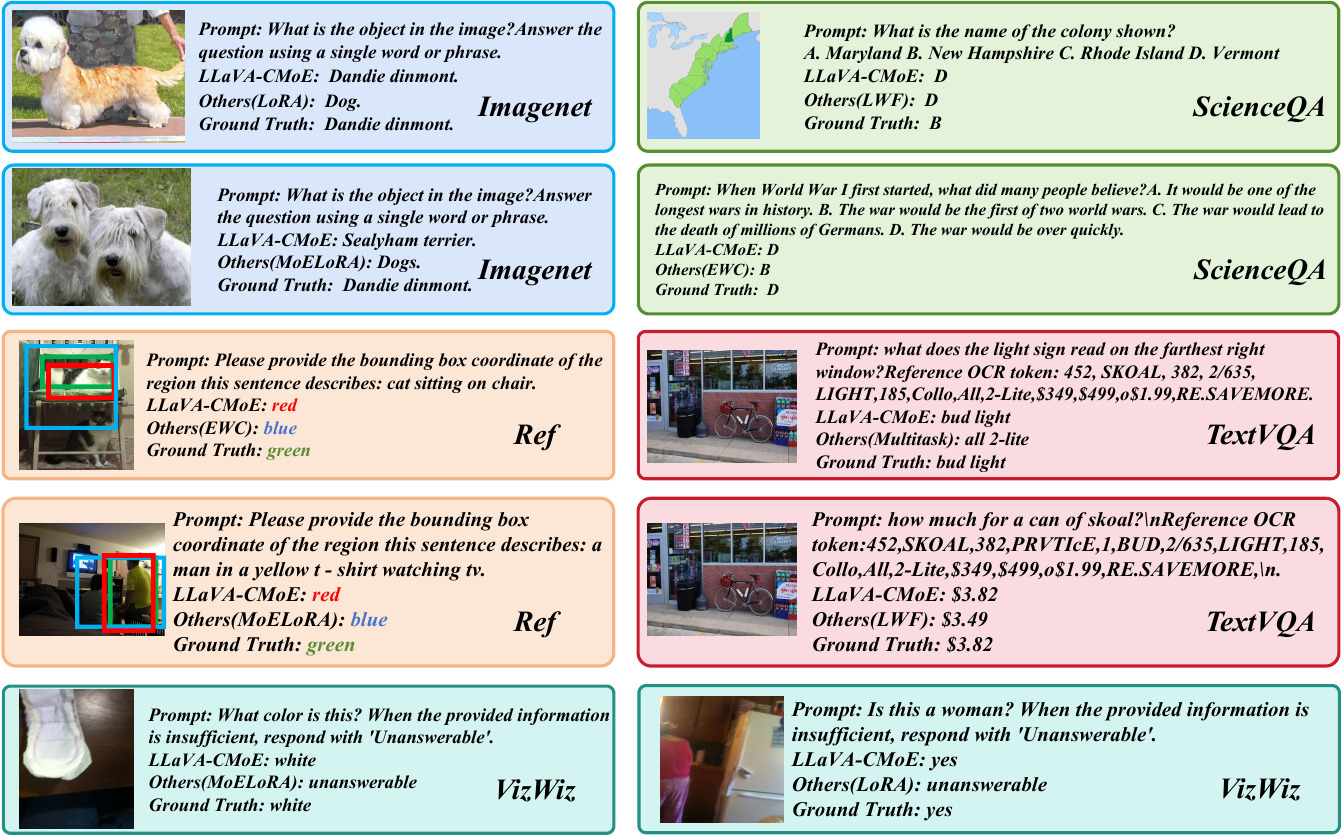}
    \caption{We randomly select five datasets, which are relatively prone to forgetting, and visualized the results of the two models for comparison.}
    \label{fig:vis_2}
\end{figure}

\paragraph{Performance over tasks.} 
We evaluate our methods' performance during the entire training process. As shown in Table~\ref{sup:tab_5}, our approach maintains excellent anti-forgetting capability after training on sequential tasks, particularly on the SQA, TQA, ImageNet, VizWiz, and Ref datasets. By comparison, PTL exhibits slightly weaker performance on the VQAv2 and GQA datasets due to a slight classification confusion issue. Nevertheless, its performance remains comparable to or better than existing methods, as demonstrated in our main manuscript.

\begin{table}[!t]
\caption{The performance on eight datasets during sequential training. Training order: SQA $ \rightarrow $ TQA $ \rightarrow $ ImageNet $ \rightarrow $ GQA $ \rightarrow $ VizWiz $ \rightarrow $ Ref $ \rightarrow $ VQAv2 $ \rightarrow $ OCR-VQA.}
\label{sup:tab_5}
\renewcommand\arraystretch{1.1}
\renewcommand\tabcolsep{2.0pt}
\centering
\resizebox{0.8\linewidth}{!}{
\begin{tabular}{lcccccccc}
\toprule
\multirow{2}{*}{\textbf{Training Dataset}} &
\multicolumn{8}{c}{\textbf{Accuracy on Each Task}} \\ \cmidrule(r){2-9} 
& SQA & TQA & ImageNet & GQA & VizWiz & Ref & VQAv2 & OCR-VQA \\ 
\midrule
 {SQA} & 79.01 & - & - & - & - & - & - & -  \\
{TQA} & 79.01 & 59.94 & - & - & - & - & - & -  \\
{ImageNet} & 79.01 & 59.71 & 96.85 & - & - & - & - & -  \\
{GQA} & 79.00 & 59.28 & 96.31 & 56.43 & - & - & - & -  \\
{VizWiz} & 78.21 & 58.59 & 95.13 & 55.86 & 57.44 & - & - & -  \\
{Ref} & 78.21 & 58.58 & 95.12 & 54.40 & 57.42 & 25.63 & - & -  \\
{VQAv2} & 77.73 & 58.31 & 94.51 & 49.86 & 56.31 & 24.02 & 65.15 & -  \\
{OCR-VQA} & 77.55 & 58.17 & 94.50 & 48.91 & 55.45 & 23.40 & 56.40 & 62.01  \\
\bottomrule
\end{tabular}
}
\vspace{3mm}
\end{table}

\subsection{Training Details}
We train the model using 8 $\times$ H20 GPUs. It takes 2 hours to train the initial eight experts, followed by 15 hours for the continual learning process. Throughout the training, we set the warmup ratio to 0.03, use the AdamW optimizer, and employ torch BF16 precision with DeepSpeed stage zero2. The rank of LoRA experts is set to 64, the rank alpha to 128, and the global batch size to 128. We assign a weight of 1e-3 to the load balancing loss of the router, the KL divergence loss, and the reconstruction loss. For training the normal experts, we use a learning rate of 2e-4, and during the probe experts training, we increase the learning rate to 3e-4.



\end{document}